\documentclass{article}
\usepackage{spconf,amsmath,graphicx}

\name{Yonatan Sverdlov, Shimon Ullman}
\address{Weizmann Institution, Computer Science and applied math}
\usepackage{hyperref}
\usepackage{url}
\usepackage{graphicx}
\usepackage{diagbox}
\usepackage[nottoc]{tocbibind}
\usepackage{comment}
\title{Efficient Rehearsal Free Zero Forgetting
    Continual Learning using Adaptive Weight Modulation}
\hypersetup{
    colorlinks=true,
    linkcolor=blue,
    filecolor=magenta,      
    urlcolor=cyan,
    pdftitle={Overleaf Example},
    pdfpagemode=FullScreen,
}

\begin{document}
%
\maketitle
\begin{abstract}
Artificial neural networks encounter a notable challenge known as continual learning, which involves acquiring knowledge of multiple tasks over an extended period. This challenge arises due to the tendency of previously learned weights to be adjusted to suit the objectives of new tasks, resulting in a phenomenon called \textit{catastrophic forgetting}.
Most approaches to this problem seek a balance between maximizing performance on the new tasks and minimizing the forgetting of previous tasks.
In contrast, our approach attempts to maximize the performance of the new task, while ensuring zero forgetting. 
This is accomplished by creating a task-specific modulation parameters for each task. Only these would be learnable parameters during learning of consecutive tasks. 
Through comprehensive experimental evaluations, our model demonstrates superior performance in acquiring and retaining novel tasks that pose difficulties for other multi-task models. This emphasizes the efficacy of our approach in preventing catastrophic forgetting while accommodating the acquisition of new tasks.

\end{abstract}

\section{Introduction}
\label{sec:intro}
Artificial neural networks face a significant challenge in continual learning, which involves learning numerous tasks over a prolonged period, due to the tendency of weights learned for prior tasks to be modified to suit the objectives of new tasks, resulting in catastrophic forgetting \cite{RobertMFrench},\cite{McCloskeyandCohen}. Although recent progress in machine learning and deep neural networks has led to impressive performance improvements in various fields, existing continual learning techniques provide limited solutions and do not ensure zero-forgetting over extended periods, for example see \cite{Measuring_Forgetting}. Most approaches attempt to balance learning and forgetting, and even minor changes to the system can result in reduced performance in new tasks.
Maintaining zero-forgetting is of utmost importance as it is unacceptable to jeopardize the knowledge gained from past tasks. This is especially critical in domains like medical image detection, where the capability to identify new diseases should not come at the expense of losing the system's proficiency in recognizing existing ones. Likewise, the potential consequences of compromising an autonomous driving system through forgetting are substantial, posing significant hazards.
While there are existing solutions for continual learning, including those that address the issue of zero forgetting, they often rely on less efficient networks such as Res-Nets (as mentioned in \cite{resnet}). However, these networks may not be suitable for deployment on small local devices like MobileNet does (cited in \cite{mobilenetv2} and \cite{mobilenetv3}). Unfortunately, the performance of such efficient and compact devices falls significantly below the optimal level in the continual setting. Given the increasing demand for both zero forgetting and efficiency, it becomes crucial to achieve progress by merging these two aspects. The challenge lies in developing methods that can simultaneously achieve both zero forgetting and efficiency, as each aspect is important on its own but combining them requires further advancements.
Our findings indicate that smaller networks have a higher tendency to experience forgetting.
In this paper, we introduce one of the earliest zero-forgetting techniques. Our approach does not damage previously learned tasks, and we outperform state-of-the-art baselines. However, as we cannot modify existing weights, which means that we need to add more parameters for each new task. We aim to keep the added parameters relatively small.
Our approach offers the advantage of not requiring specific bounds on the number of tasks that can be added. We begin with a pre-trained network and expand it as necessary, ensuring that the model is appropriately sized for the tasks encountered thus far. This differs from many current approaches that use a fixed model, which can often start too large and eventually become too small to handle new tasks.
There are three main methods used in current approaches. The first method involves making sure that partial data from all tasks are available during training. This is done by interleaving data from multiple tasks during learning, which helps to prevent large amounts of forgetting if the network weights can be optimized for performance on all tasks simultaneously. However, this approach, known as Rehearsal Methods, is often impractical for learning large numbers of tasks, such as in our setting, as it would require storing and replaying data that is proportional to the number of tasks. Additionally, data may be missing due to privacy concerns, as in the case of medical imaging.
\newline
The second main approach proposed is regularization-based methods, where the pre-trained model on old tasks is given and the model attempts to learn the new objective while minimizing changes in the relevant existing weights to avoid forgetting. However, the main issue with regularization-based methods is that they cannot guarantee low forgetting, and in long continual scenarios, the accuracy of old tasks often decreases dramatically, resulting in poor learning of new tasks as well, as shown in several studies such as \cite{Measuring_Forgetting}, \cite{PLOP}, and \cite{zhang2022representation}.
\newline
Architectural methods are another approach to mitigate catastrophic interference in continual learning. However, they typically require a drastic increase in computational and storage requirements, often doubling the model parameters for each new task \cite{pathnet}, \cite{pnn}, \cite{extension}. Task masking \cite{piggyback} is one of the single efficient zero forgetting methods, where a single bit is learned per weight, and many studies have investigated its convergence and stability  \cite{lottery_ticket1}, \cite{lottery_ticket2}, \cite{lottery_ticket3}, \cite{lottery_ticket4}.
For over-parameterized networks, like ResNet50 \cite{resnet}, masking shows almost optimal performance as was shown in \cite{piggyback}, but as we will show for much more efficient architecture with similar performance used by computation-bounded devices like Mobile Net \cite{mobilenetv2}, the performance of Masking is far from optimal. Our proposed adaptive method shows much better performance than masking, with the same parameter addition, and parameters can be reduced while still outperforming masking.
\newline
\textbf{Our contributions are as follows:}
\begin{itemize}
  \setlength\itemsep{0em}
  \item We present a new method with guaranteed zero forgetting for any number of added tasks.
  The method reaches high accuracy in different domain classification tasks, outperforming existing state-of-the-art methods.
  \item The method adds a set of parameters using task modulation for each task. Our method's total number of parameters is still the same or lower than the alternative methods in all our experiments.
    The method can continue adding any number of tasks without inner-task interference.
  \item The algorithm is faster and requires less storage than alternatives.
  \item Rehearsal free and no need of storing old parameters.
\end{itemize}
\section{Related Work}
\subsection{Multi task learning}
Multi-task learning (MTL) involves training a shared model to perform multiple tasks. This approach has several benefits, such as increased data efficiency, reduced over-fitting through shared representations, lower computation and resource requirements, and faster learning by utilizing auxiliary information. For a thorough overview of the field, refer to \cite{MTL-servey}. MTL can be classified into two primary settings: joint learning and continual learning.
\newline
\textbf{Joint learning}
The objective is to acquire proficiency in all assignments by considering the availability of data from all tasks during training. The key benefit of this approach is the utilization of extensively shared representations for all tasks, as observed in the case of learning surface normal estimation and depth estimation \cite{Geonet}, semantic segmentation and depth estimation \cite{segmenation_and_depth}. Nevertheless, this approach is often impractical due to constraints such as data storage limits and the real-time requirements of systems.
\newline
\textbf{Continual learning}
In Continual learning, access to only the data from the current task is available. The principal challenge of this learning approach is to reduce catastrophic forgetting, i.e., the problem of learning new tasks without erasing knowledge acquired from previous tasks. When the tasks being learned are independent, the learning process is analogous to training a distinct network for each task. Multi-task learning (MTL) approaches assume that the tasks are not independent, and aim to exploit this property to add new tasks efficiently without forgetting previous knowledge. However, some scenarios may be hybrid, where a small amount of data from previous tasks is available, as in \cite{gem} and \cite{rieman_walk}. Continual learning can be categorized into two main scenarios: task incremental and class incremental. In task incremental, the task IDs are known during evaluation time, whereas in class incremental, the model should differentiate between all tasks that have been seen so far, even if their IDs are unknown. In our paper, we focus on task incremental problems.
\subsection{catastrophic forgetting}
\label{sec:CF} 
Catastrophic forgetting has been a well-known phenomenon in continual learning since the $1980s$, as evidenced by early research such as \cite{RobertMFrench}, \cite{McCloskeyandCohen}, and \cite{cite_first_papers_about_forgetting}. In contrast to artificial neural networks, humans and other animals appear to be capable of learning continually, as described in \cite{EWC}. Despite the decades of research since the discovery of catastrophic forgetting, the available approaches to continual learning still offer limited solutions, as elaborated in the subsequent sections.
As an example, in \cite{Measuring_Forgetting}, various existing methods from different approaches were tested on multiple tasks. The evaluation was conducted on the CUB$200$ \cite{cub1}, \cite{cub2} dataset, which includes $200$, classification classes. The experiment began by training on $100$ classes and then incrementally adding one class at a time. Next, the remaining $100$ tasks were trained using each selected representing method. The results indicated that while offline learning achieved an overall accuracy of $62\%$, the best-performing method (Rehearsing method) demonstrated a learning of $35\%$, revealing a significant disparity. The principal challenge of continual learning is to adjust the weights to meet the new objective without significantly modifying the pertinent weights in the previous tasks. The primary methods employed to address this challenge, as detailed in \cite{Measuring_Forgetting}, include Rehearsal Methods, Architectural Methods, and Regularization Methods.
\subsection{Architectural methods}
\label{architec}
Architectural methods aim to mitigate catastrophic interference by separating what is being learned from what has already been learned. One of the earliest architectural methods is branching, also known as transfer learning. This approach involves training a network and subsequently adding another small network to the pre-trained one, with only the added network being learned.
The primary drawback of the branching approach is that the bottom features of the new model cannot be altered and can only be adjusted by modifying the new weights. This limitation can negatively impact the learning accuracy of the new tasks, leading to a relative decline in performance, as evidenced by our experiments. It is worth noting that in certain cases, adding only a classifier may result in a high level of performance (e.g. see Flowers \cite{flowers}, in \autoref{baseline}). In contrast, our proposed method involves modulating all body model weights, affecting the model from the low-level features to the high-level ones.
The branching approach has found extensive use in various fields such as NLP \cite{transferNLP}, \cite{sun2019fine}, VLP (vision language pretraining) \cite{chen2022vlp}, and image classification \cite{transfer}. In VLP, new tasks are introduced by performing fine-tuning on the pre-trained model. While this approach can often accommodate new tasks, it has limitations and can lead to significant forgetting. To mitigate the effects of forgetting, new variations of fine-tuning such as LORA \cite{LORA} are being developed. The primary advantage of this approach is that it requires training only a small number of new parameters. Dual-weight methods \cite{Measuring_Forgetting} maintain a set of weights for a single-network architecture for each task.
PNN (progressive neural network) and ENN (Extended Neural Networks) are noteworthy instances that tackle the issue of forgetting by employing individual networks for each task. These approaches incorporate the previously learned features into the current layer through a lateral connection originating from the previous layer (or by employing a learnable sum of kernels, as in ENN). PNN and ENN offer the benefit of feature sharing and help prevent overfitting. However, a drawback of these methods is that they necessitate the creation of a new network for each additional task.
Pack-Net \cite{PackNet} is a technique that draws inspiration from network pruning methods to exploit redundancies in deep networks, freeing up parameters that can then be used to learn new tasks. The process involves iterative pruning and network re-training to sequentially incorporate multiple tasks into a single network. After learning a task, a task-specific sub-network is extracted by pruning the network and frozen for the subsequent tasks. When selecting the task-specific sub-network, each task builds on the sub-network corresponding to the previous task with some extensions based on the pruning outcomes. However, the number of tasks that can be learned with this approach is very limited, as no free weights remain for training after a certain number of tasks, as shown in \cite{piggyback}.
The primary zero-forgetting architectural method is weight masking \cite{piggyback} which serves as the main zero-forgetting method. In weight masking, a binary mask is learned for each weight, allowing for the addition of only a percent of one over the machine precision, typically $16$ or $32$, resulting in a relatively small addition. Weight masking leverages the over-parameterization of large models like Res Nets, but its performance is sub-optimal in compact versions, such as Mobile-Net \cite{mobilenetv2}, as we will show. The primary drawback of masking is that it's not adaptive, and it cannot scale up for improved performance by adding more parameters. Our solution is an adaptive architectural method.
An alternative approach to weight modulation is \textbf{neural} modulation, as described in the paper cited as \cite{neural_mod}. Neural modulation involves modulating neurons either through binary masks or directly using learnable parameters. While the paper successfully applied binary neuron modulation to solve simple tasks, it remains unclear how this method can be adapted to large models. In large models, the number of neurons is negligible compared to the number of parameters (for instance, in ResNet18, it's less than 0.1\%). Consequently, it becomes challenging to achieve meaningful changes using neural modulation in such cases. While neural modulation is a promising zero-forgetting scheme, its applicability to real-world large-scale applications is limited.
\subsection{Rehearsal methods}
The primary objective revolves around the retention of pertinent information from previous tasks, which is then amalgamated with fresh data to prevent catastrophic forgetting regarding the prior tasks. In the study referenced as \cite{gdumb}, the data from prior experiments undergoes shuffling alongside the new data, subsequently facilitating optimization of the model on both datasets. Conversely, certain studies such as \cite{gem} opt to aggregate gradients from all tasks into a novel gradient that is employed across all tasks.
However, there are notable drawbacks to this approach. One major limitation is the linear growth of storage requirements with the increasing number of tasks, making it impractical for scenarios involving a large number of tasks over an extended period. Additionally, this solution often falls short when only a limited amount of data can be stored, as evidenced by the significant decline in performance observed in \cite{Measuring_Forgetting}.
Additionally, privacy concerns, particularly in domains like medical imaging, impose restrictions on the utilization of past samples. Due to privacy regulations and ethical considerations, access to and storage of patient data becomes limited, making it impractical or even prohibited to use previously collected samples for continual learning purposes. This poses a significant challenge in devising effective solutions for zero-forgetting in scenarios where data privacy is of utmost importance.
Therefore, this approach serves as a compromise between storing all data to ensure optimal performance and completely omitting storage, which leads to significant forgetting. It should be noted that our proposed solution eliminates the need for data rehearsal, which can be impractical and inefficient in certain cases.
Another significant drawback is the performance impact when the number of classes or samples per class in the new task differs greatly from the original task. In such cases, data rehearsal methods can result in severe forgetting or low performance on the new task, as the model attempts to adapt to the task with more samples or classes. This renders rehearsal methods ineffective in scenarios where there is a substantial discrepancy between the original and new tasks.
Furthermore, it is important to acknowledge that zero forgetting is not guaranteed with this approach, especially when the number of stored samples is low. The potential for forgetting remains a concern, and the effectiveness of the solution depends on the quantity and quality of the stored samples.
\subsection{Regularization methods}      
In these methods, a regularization factor is introduced to the objective that measures the dissimilarity between the current and previous weights. More specifically, given the objective function $L_{B}$ for the current task, a regularization function $R_{i,\theta}$ is added to the objective that may depend on the current parameters, step $i$, and previous model weights $\theta^{*}$. The resulting objective for the new task is $L_{new}(\theta) = L_{B}(\theta) + \lambda \cdot R_{i,\theta}(\theta^{*},\theta)$, where $\lambda$ is a hyper-parameter that controls the regularization strength. This approach aims to strike a balance between preserving the old tasks' knowledge by imposing a high regularization factor and allowing for the model's flexibility to learn the new task by using a low regularization factor, which could result in forgetting the old tasks.
Regularization methods exhibit various significant shortcomings that hinder their efficacy in mitigating catastrophic forgetting. One notable concern is the potential accumulation of errors over prolonged periods of continual learning, resulting in substantial levels of forgetting pertaining to previously learned tasks (see \autoref{fig:LWF} in our results). Even cutting-edge image segmentation techniques, as exemplified in \cite{PLOP} and \cite{zhang2022representation}, which demonstrate proficiency in few-task scenarios, exhibit subpar performance in the context of extended continual task challenges where multiple tasks are sequentially acquired.
Furthermore, the computation of regularization factors often necessitates iterating over all parameters during each training iteration. This requirement leads to protracted training times and extensive computational demands, as evidenced by studies such as \cite{EWC}, \cite{synaptic_intelligence}, \cite{MAS}, and \cite{rieman_walk}. Consequently, this approach may become impractical for large-scale models and complex tasks.
Moreover, the storage of previous weights within the model doubles the parameter count and memory utilization, thereby further complicating the training process. Collectively, these limitations impede the effectiveness of regularization methods in numerous practice settings.
\section{Problem statement and setup}
\subsection{Goal}
The task involves multiple tasks $T_{i}$ and their corresponding data sets $D_{i}$, where $i \in [N]$ and $N$ represent the total number of tasks. The objective is to learn all tasks accurately with minimal forgetting while ensuring that the model is only trained on $T_{i}$ with access to $D_{i}$ only. This creates a trade-off between high learning and low forgetting, but our specific goal is to ensure zero forgetting while maximizing task accuracy.
\subsection{Our method}
\begin{figure}[htp]
    \hspace*{-0.2cm}   
    \includegraphics[width=9cm]{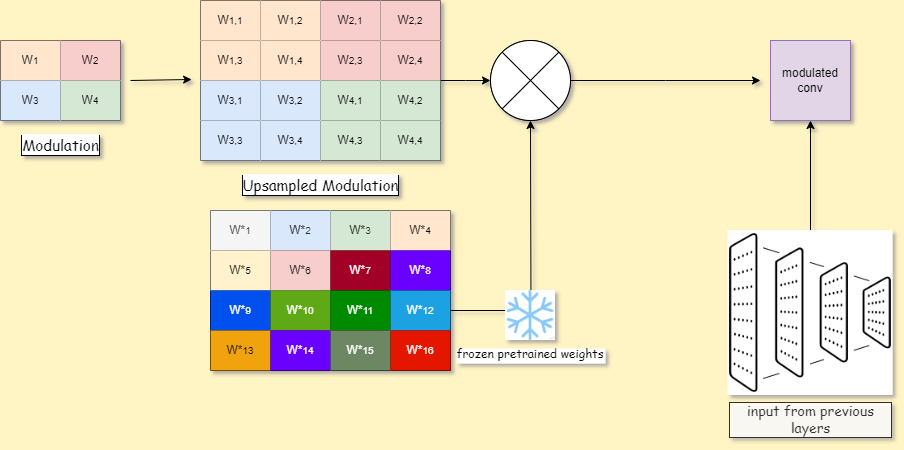}
    \caption{An overview of the modulation $(m_1, m_2) = (2, 2)$. A similar operation is being done to all the network layers.}
    \label{fig:modaultion_example}
\end{figure}

The objective of our approach is to effectively adjust the weights of a pre-trained neural network by introducing learnable parameters. These learnable parameters, called task modulation parameters, are specific to each task.
We focus on modulating convolution layers, which have a shape defined by the number of input channels ($C_{in}$), the number of output channels ($C_{out}$), and the kernel size ($k_{1}, k_{2}$). To achieve modulation, we utilize modulation resolution parameters, denoted as $m_{1}$ and $m_{2}$, which determine the resolution of the modulation.
To enable modulation, we generate a learnable parameter called $Mod$, which has a shape of $(\frac{C_{in}}{m_{1}}, \frac{C_{out}}{m_{2}}, k_{1}, k_{2})$.
We initialize Mod around one.
The values of $m_{1}$ and $m_{2}$ are chosen such that they are smaller than the original $C_{in}$ and $C_{out}$ values, respectively. This learnable parameter is then expanded to match the shape of the original weights using interpolation, typically employing a technique like bi-cubic interpolation.
The modulation is applied by performing an element-wise multiplication between the original weights ($W$) and the modulation parameter ($Mod$), resulting in an updated set of weights ($W := W \cdot Mod$). After modulation, the usual convolution operation is performed using these updated weights.
This modulation approach can also be extended to linear layers by applying the modulation parameter to the corresponding weight matrix.
An illustrative example of $(2,2)$ fully connected modulation is presented in \autoref{fig:modaultion_example}
For each task, we select a fixed resolution that is used consistently across the entire network. By default, we use a resolution of $(4,4)$ or $(2,8)$. However, this resolution can be adjusted to find the optimal balance between accuracy and the number of parameters, as discussed in the context of 
\autoref{baseline}).


This introduces a novel trade-off between accuracy and the number of added parameters, in contrast to regularization methods which focus on the trade-off between old and new task accuracy.
\subsection{Training \& evaluation}
Our approach starts with a pre-trained Mobile-Net V2 with 3.5M parameters, \cite{mobilenetv2} on Image-Net \cite{ImageNet}, which is widely used in real-world scenarios and is more efficient and compact than other models like ResNet. 
For each new task, we choose the modulation ratio and is added to the base network and add a linear classifier at the top of the model and we train the task parameters including the classifier while freezing all other weights. To address the issue of batch normalization (BN) layers, which can have running statistics ($\mu$, $\sigma$) that are shifted toward new tasks, we store the normalization statistics at the end of training for each individual task and use them during evaluation, as was done in \cite{piggyback}. This approach ensures zero forgetting. When evaluating the model, we are given an image and task specification, which the model uses as inputs along with the stored BN statistics and learned modulation.
When we compare our methods to alternatives, we include all the added parameters above in counting the number of parameters.
\section{Experiments}
\subsection{Implementation Details}
\textbf{Data sets:}
\newline
In our study, we used the dataset introduced in \cite{piggyback}, which includes Flowers \cite{flowers}, Stanford Cars \cite{stanford_cars}, CUBS \cite{cub1, cub2}, Sketches \cite{sketch}, Wiki-Art \cite{WikiArt}, and we additionally included the Food$101$ dataset \cite{food101}. The number of samples per dataset can be found in \autoref{samples}.
Wiki-Art and Food101 are considered out-of-distribution and the most interesting datasets in our study as they are different from Image-Net and we expect a high difference between different methods.
\newline
\textbf{Architecture Details:}
We used the Mobile-Net V2 pre-trained on Image-net with a learning accuracy of 71.89\% architecture due to its comparable performance to ResNet$50$ but with a significantly smaller size( By a factor of about 8).
Our results, as presented in \autoref{accuracy}, indicate that both ResNet$50$ and the more compact Mobil-Net V$2$ have similar optimal performance on the new datasets, supporting the choice of Mobile-Net due to its smaller size. 
To ensure fairness in comparisons to our alternatives, we compare models with an equal number of parameters, measured in total storage requirements. For example in comparing our method to the masking method \autoref{accuracy}, we use machine precision of $16$ bits, and a resolution of $(4,4)$, or $(2,8)$ which makes the model equivalent to the masking that uses a single bit per weight.
We used the $16$ bit precision throughout since we found it has only a minor effect on accuracy with compared with the higher 32 precision.
\newline
\begin{table*}[hbt!]
\centering
\small
\begin{tabular}{|c|c|c|c|}
 \hline
 Data-set & Mobile-Net V2 & ResNet50 \\
 \hline
 Stanford-Cars & 87.19\% & 91.7\% \\
  \hline
  Flowers  & 91.3\% & 95\%  \\
  \hline
  CUBS200 & 80.53\% & 82.83\% \\
  \hline
  Sketches & 78.37\%  & 80\% \\
  \hline
  Wiki-art & 68.82\% & 75\%  \\
  \hline
 \end{tabular}
\caption{Mobile-Net vs ResNet50 performance on several data-sets.}
\label{accuracy}
\end{table*}

\textbf{Hyper-parameters and training}
We repeated each experiment five times and recorded the average accuracy. The Adam optimizer was used for training, with a learning rate decay of $0.1$ after $15$ epochs. We conducted a brute force search across learning rates $lr \in [1e-1, 1e-2, 1e-3, 1e-4]$. The training was carried out for $30$ epochs.
\begin{table*}[hbt!]
\centering
\small
\begin{tabular}{|c|c|c|c|}
 \hline
 Data-set & Train & Test & Classes \\
 \hline
 Image-Net & 1,281,144 & 50,000 & 1,000 \\
  \hline
 Stanford-Cars & 8,144 & 8,041 & 196 \\
  \hline
  Flowers  & 2,040 & 6,149 & 102 \\
  \hline
  CUBS200 & 5,994 & 5,794 & 200 \\
  \hline
  Sketches & 16,000 & 4,000 & 250 \\
  \hline
  Wiki-art & 42,129 & 10,628 & 195 \\
   \hline
  Food101  & 75,500 &  25,250 & 101  \\
  \hline
 \end{tabular}
\caption{Summary of used datasets.}
\label{samples}
\end{table*}
\subsection{Results}
We evaluate our method by contrasting it with two baseline techniques: the masking method and LwF. The masking method closely resembles our approach in terms of utilizing task modulation and successfully achieving zero forgetting. On the other hand, the other zero forgetting models, namely Pack-Net and neural modulation, fall short in attaining state-of-the-art outcomes and are not adaptable to models and data that are of high scale and resolution. As for LwF, it stands out as the most widely used regularization method that balances accuracy and forgetting.
Data rehearsal methods are not commonly employed due to challenges such as class imbalance and the limited number of samples per class.
In addition, we also compare the impact of two different basic training approaches: solely training the weight of the final classification layer (transfer learning) and training all model weights (full model training). This comparison sheds light on the efficacy of these distinct training strategies.
In \autoref{accuracy}, we present the learning accuracy achieved by each baseline model for every dataset in our evaluation. All the presented results are on the test dataset.
\subsection{Comparison to Masking}
\label{sec: baselines}
As outlined in the relevant literature (see \autoref{architec}), the masking approach involves learning additional binary weights for each task. These learned weights have a size of 1 divided by the digital precision of the model weights, and in our experiments, we set the precision to 16. Despite this adjustment, The effect on the performance of the pre-trained network is negligible, with only a marginal $0.27\%$ reduction noted (from an initial accuracy of $72.15\%$ to $71.88\%$).
To ensure an equivalent number of parameters, each modulation $(m_{1},m_{2})$ resolution must hold $m_{1} \cdot m_{2} \geq 16$. 
In \autoref{resolution}, we show for each dataset the modulation resolution and the interpolation method.

\begin{table*}[hbt!]
\centering
\small
\begin{tabular}{|c|c|c|c|}
 \hline
 Data-set & Resolution & Interpolation \\
 \hline
 Stanford-Cars & (2,8) & nearest  \\
  \hline
  Flowers  & (2,8) & bi-cubic \\
  \hline
  CUBS200 & (2,8) & bi-cubic \\
  \hline
  Sketches & (2,8) & nearest-exact \\
  \hline
  Wiki-art & (4,4) & bi-cubic  \\
   \hline
  Food101  & (4,4) &  bi-cubic   \\
  \hline
 \end{tabular}
\caption{Modulation resolution and interpolation method.}
\label{resolution}
\end{table*}

The results are present in \autoref{baseline}
Remarkably, our results on the WikiArt and Food101 datasets demonstrate superior performance compared to masking. This discrepancy can be attributed to the fact that the other datasets, similar to Image-Net, exhibit significant class overlap, while WikiArt and Food101 contain classes not found in Image-Net. In addition in Flowers and CUBS read-out alone shows very good performance, indicating the high similarity to Image-Net). Consequently, these datasets(WikiArt and Food101) differ substantially from Image-Net, and the binary mask alone is insufficient for effective transfer. However, on the other datasets, the performance is comparable to masking.
Overall, the modulation approach achieves a performance improvement of 2.6\% over masking. Additionally, we provide the average performance across all datasets.
\newline
\begin{table*}[hbt!]
\hskip 2.5cm
\begin{tabular}{|c|c|c|c|c|c|c|c|}
\hline
\backslashbox{Method/}{Performance}
&\makebox[1.5em]{Cars}&\makebox[2.5em]{Flowers}&\makebox[2em]{CUBS}
&\makebox[3em]{Sketches}&\makebox[3em]{WikiArt}&\makebox[1.5em]{Food}&\makebox[2em]{Avg} \\\hline\hline
Transfer learning (\cite{transfer})  & 52.2 & 88.42 & 60.7 & 64.13 & 55.43 & 63.67 & 66.11 \\\hline
Piggyback (\cite{piggyback}) & \textbf{84.85}  & \textbf{91.94} & \textbf{71.0} & \textbf{78.4} & 55.98 & 73.1 & 76.05   \\\hline
\textit{ZFCL} (ours) &84.0 & 90.35 & 72.9 & 77.93 & \textbf{69} & \textbf{78.5} 
& \textbf{78.66} \\\hline
\hline
Full-model training & 90.57 & 92.42 & 79.8 & 79.35 & 74.44 & 83.4 & 82.74 \\\hline
\end{tabular}
\label{baseline}
\caption{We investigate several methods for learning multiple datasets, which include weight masking, transfer learning, fine-tuning, and weight modulation. The results of our experiments are reported in percentages.
}
\end{table*}

\subsection{Comparison to regularization}
\begin{figure}[htp]
    \centering
    \hspace*{-1cm}   
    \includegraphics[width=10cm]{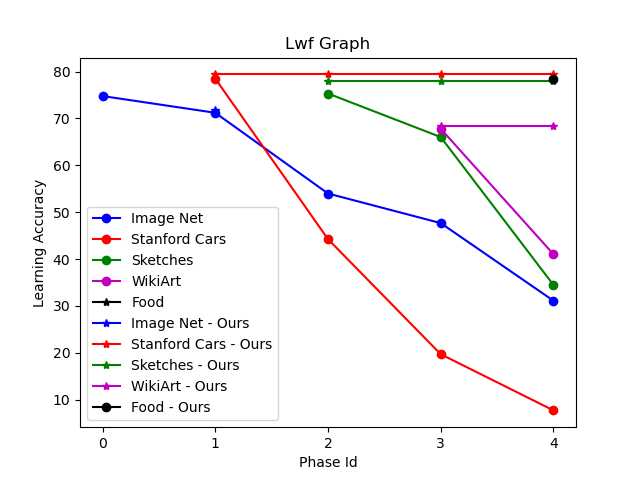}
    \caption{\textbf{The performance of LwF and ours in sequential task addition.} We start(for both ours and LwF) with a pre-trained network, and at each time point, we train for additional datasets.
    upon sequential adding of tasks, LwF performance drops significantly on the older tasks. Ours however remains high without change.
    Upon observation, it becomes evident that although LwF is trained to align with our learning, the decline in performance is significant, marked by a drastic and dramatic drop.}
    \label{fig:LWF}
\end{figure}

In order to ensure a fair comparison, we conducted training on MobileNet V3 \cite{mobilenetv3}, which had previously been pre-trained on Image-Net with a learning accuracy of 75.9\% and consisted of $5.5$ million parameters. It is important to note that the parameter count of this model exceeded that of our modified MobileNet V2 \cite{mobilenetv2}, which included an additional task modulation and had $4.8$ million parameters.
To prevent any distribution shift in the batch normalization layers, we trained the network with these layers in evaluation mode. This approach ensured that the running statistics remained unchanged and were not influenced by the specific batch being processed.
The regularization network was trained sequentially on the Stanford Cars, Sketches, Wiki-Art, and Food101 datasets. For each dataset, we performed a hyper-parameter search to determine the appropriate regularization factor that would match the accuracy achieved by our model. For instance, on the Stanford Cars dataset, we used a factor that yielded an accuracy of 78.5\%.
We assessed the performance of the network on all the tasks that were previously learned. At each step, we measured the accuracy achieved on the newly learned dataset, as well as the residual learning accuracy on all the tasks learned earlier. This methodology enabled us to evaluate the impact of the regularization process on the overall performance.
The findings (see \autoref{fig:LWF}) unequivocally indicate that the Learning without Forgetting (LwF) method experienced a notable decrease in accuracy across various datasets, namely Image-Net, Stanford Cars, Sketches, and WikiArt. Specifically, there was a 40\% drop in accuracy observed in Image-Net, a 70\% drop in Stanford Cars, a 40\% drop in Sketches, and a 27\% drop in WikiArt.
This highlights the importance of our zero-forgetting method, which effectively prevents any potential sequential forgetting of previously learned tasks.
In \autoref{lwf}, we show the learning accuracy of each dataset, at each phase step.
\begin{table*}
\centering
\begin{tabular}{|c|c|c|c|c|c|}
 \hline
 Phase & Image-Net & Stanford Cars & Sketches & Wiki-Art & Food \\
 \hline
 Phase 1 & 74.78 & & &  &  \\
 \hline
 Phase 2 & 71.21  & 78.5 &  & &  \\
 \hline
 Phase 3 & 53.98  & 44.24 & 75.32 & &  \\
\hline
Phase 4 & 47.67 & 19.69 & 66.025 & 67.8 & \\
\hline
Phase5 & 31.17 & 7.7 & 34.47 & 41.02 & 78.51    \\ 
\hline
\textit{ZFCL}(Ours) & \textbf{71.88} & \textbf{78.5} & \textbf{77.925} & \textbf{ 68.3} & \textbf{78.5}   \\
\hline
 \end{tabular}
 \caption{LwF learning, we employ the Learning without Forgetting (LwF) method \cite{learning_without_forgetting} to learn multiple datasets. The results, reported in percentages, demonstrate the learning accuracy achieved using the LwF approach for each dataset.}
 \label{lwf}
\end{table*}

\subsection{Adaptive modulation}
\begin{figure}[htp]
    \hspace*{-0.7cm}   
    \includegraphics[width=10cm]{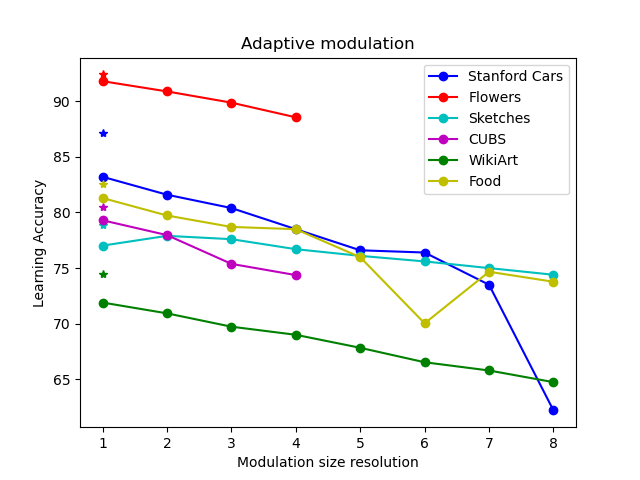}
    \caption{\textbf{Modulation size and its corresponding accuracy.} 
    We present the relationship between modulation size and its corresponding learning accuracy. Additionally, we include the results of full model training, denoted by the * symbol.
    Upon analyzing multiple datasets, we have discovered that reducing the modulation size in our adaptive modulation approach has minimal effect on the learning process. Initially, the full model achieves successful learning, highlighting the presence of an optimization gap between training all model weights and training with a resolution modulation of $(1,1)$.}
    \label{fig:AM}
\end{figure}

Our approach introduces a significant innovation by incorporating modulation with varying resolution sizes. This is achieved through an exhaustive search process, where we explore a range of resolutions, specifically focusing on sizes $(i, i)$ where $2 \leq i \leq 8$. Through this systematic investigation, we are able to determine the optimal configuration for our modulation method.
The search for an optimal resolution gap revealed an optimization problem at the highest resolution( shown in star symbols in \autoref{fig:AM}). Using a resolution of $(1,1)$ should be equivalent to training the full model without using any resolution parameters.
However, the results show a consistent optimization gap between training all the model weights and training modulation of the shape $(1,1)$. This discrepancy indicates that training a modulation of shape $(1,1)$ leads to different optimal weights with lower test accuracy. A potential reason behind these results is overfitting which is supported by the nearly perfect accuracy achieved on the training dataset. 
Results for each dataset and modulation resolution can be found in \autoref{adaptive}.
\begin{table*}[hbt!]
\centering
\footnotesize
\begin{tabular}{|c|c|c|c|c|c|c|c|c|}
 \hline
 Data-set name & (1,1) & (2,2) & (3,3) & (4,4) & (5,5) & (6,6) & (7,7) & (8,8) \\
 \hline
 Stanford-Cars & 87.4\% & 85.5\% & 84.7\% & 84.0 \% & 81.4 \% & 81.4\% & 75.1 \% & 77.9\%  \\
  \hline
  Flowers  & 90.89\% & 89.88\% & 88.56\% & 86.25\% & 87.67\% & 82.35\% & 84.69\%  \\
  \hline
  CUBS200 & 77.967\%  & 75.38\% & 74.37\% & 70.25\% & 69.76\% & 61\% & 71.2\%  \\
  \hline
  Sketches & 77.9\% & 77.6\% & 76.7\% & 76.1\% & 75.6\% & 75.0\% & 74.4\%  \\
  \hline
  Wiki-art & 70.93\% &  69.73\% & 69\% & 67.83\% & 66.53\% & 65.79\% & 64.74\%    \\ 
   \hline
  Food101  & 79.73\% & 78.7\% & 78.5\% & 75.98\% & 70.03\% & 74.67\% & 73.77\%  \\
  \hline
 \end{tabular}
 \caption{Adaptive modulation, learning accuracy of several modulation resolutions.}
 \label{adaptive}
\centering
\small
\end{table*}

\section{Discussion}
In our research, we have introduced a new approach called 
\textit{ZFCL} that effectively tackles the difficulties of continual learning and catastrophic forgetting. Unlike most existing models, our approach successfully overcomes the usual trade-off between accuracy and forgetting, and it exhibits outstanding performance when faced with unfamiliar tasks. Our model stands out due to its unique features, which are not commonly found in other continual learning models. Firstly, we utilize task modulation, a mechanism where synapses are specifically learned and utilized for each task. This characteristic bears resemblance to the brain's cortical circuitry observed in previous studies (\cite{EWC}, \cite{shimon1}, \cite{shimon2}). As a result, our model has a maximum modulation size, yet it can also adapt to learning new tasks with partial modulation. This indicates the potential to optimize the model's ability to learn new tasks without excessively increasing the number of additional weights needed.
\bibliographystyle{IEEEbib}
\bibliography{paper,refs}
\end{document}